\documentclass[preprint,12pt]{elsarticle}

\usepackage{amsmath,amssymb,amsfonts}
\usepackage{algorithm}
\usepackage{algorithmic}
\usepackage{graphicx}
\usepackage{textcomp}
\usepackage{xcolor}
\usepackage{xurl}
\usepackage{hyperref}
\usepackage{tikz}
\usetikzlibrary{positioning,fit,backgrounds,calc,arrows.meta}

\journal{Journal Name}

\begin{document}

\begin{frontmatter}

\title{A Controlled Benchmark of Quantum-Latent GAN Augmentation for Brain MRI}

\author[inst1]{Syed Mujtaba Haider}
\affiliation[inst1]{organization={Department of Mathematics},
            addressline={University of Pavia},
            city={Pavia},
            country={Italy}}
\author[inst1]{Silvia Figini}
\affiliation[inst1]{organization={Department of Political and Social Sciences},
            addressline={University of Pavia},
            city={Pavia},
            country={Italy}}
\begin{abstract}
Medical image classification is often constrained by limited labeled
data, motivating generative augmentation; recently, quantum generative
models have been proposed for this purpose, frequently reporting accuracy
gains. However, such claims are typically based on single training runs,
do not match the parameter budgets of the quantum and classical
generators, and do not characterize the data regime in which any benefit
appears. We present a controlled benchmark that isolates the contribution
of a quantum generator to brain-MRI augmentation. Images are encoded into
a KL-regularized latent space in which a conditional Wasserstein GAN with
gradient penalty is trained using either a variational quantum generator
or a classical generator of near-identical parameter count (1648 vs.\
1632). Synthetic samples are decoded and used to augment a pretrained
classifier across labeled-data fractions from 5\% to 100\%, evaluated over
eight random seeds with paired significance testing (with multiple comparison
correction) and with intra-set diversity and latent-distribution analyses.
Across all fractions, no augmentation variant significantly outperforms
real data only training, and the quantum and classical generators are
statistically indistinguishable. Any low-data benefit behaves as
regularization rather than faithful data expansion: synthetic samples are
off distribution and severely mode-collapsed precisely where data is
scarce, and the quantum generator is no more diverse than its classical
counterpart. We release the protocol as a testbed for rigorous evaluation
of quantum generative augmentation in medical imaging.
\end{abstract}


\end{frontmatter}

\section{Introduction}

Deep neural networks are widely being used  for brain-tumor classification from magnetic resonance imaging (MRI), but their performance relies on the availability of large, balanced and labeled datasets \cite{al2023brain}. In clinical applications such datasets
are costly to annotate, unevenly distributed across diagnostic
categories, and often restricted by privacy constraints. Data
augmentation is the usual remedy: beyond standard geometric transforms,
generative adversarial networks (GANs) have been widely used to synthesize
additional training images and to mitigate class imbalance in medical
imaging \cite{guo2023medgan, makhlouf2023use}.

More recently, quantum generative models have been proposed for the same
purpose. Quantum GANs (QGANs) and related variational circuits are
attractive on near-term (NISQ) hardware because a small number of
parameterized qubits can, in principle, express rich distributions, and a
growing body of work reports accuracy improvements when quantum-generated
samples augment classical classifiers \cite{chang2024latent, tsang2023hybrid}.
Because directly generating high-resolution images with shallow circuits
is impractical, several approaches operate in a learned latent space,
generating compact codes that are subsequently decoded into images.

Despite these encouraging reports, the evidence that the \emph{quantum}
component is responsible for any observed benefit is weak. Existing studies
typically (i) report results from a single training run, without seeds or
confidence intervals; (ii) compare a quantum generator against a classical
baseline of a different parameter count, so improvements cannot be
attributed to quantum structure rather than capacity; (iii) evaluate
augmentation at a single dataset size, leaving unclear the regime in which
it actually helps; and (iv) report downstream accuracy without
characterizing the quality or diversity of the generated samples. As a
result, a basic question remains unanswered: does a quantum latent
generator contribute anything beyond a parameter-matched classical
generator for medical image augmentation, and if so, under what conditions?

This paper addresses that question with a controlled benchmark. Brain-MRI
images are encoded into a KL-regularized latent space, in which a
conditional Wasserstein GAN with gradient penalty is trained using either a
variational quantum generator or a classical generator of near-identical
parameter count. Synthetic samples are decoded and used to augment a
pretrained classifier across labeled-data fractions ranging from 5\% to
100\%. We evaluate over eight random seeds with paired significance
testing, and we analyze generation quality through intra-set diversity
metrics and latent-space distribution overlap rather than downstream
accuracy alone.

Our results indicate a parity outcome. Across every data fraction, no
augmentation variant significantly outperforms real-data-only training,
and the quantum and classical generators are statistically
indistinguishable at a matched parameter budget. The modest gains that
appear in the low data regime behave as regularization rather than
faithful data expansion: the generated samples are off-distribution and
severely mode collapsed precisely where labeled data is scarce. These findings suggest that previously reported
quantum augmentation advantages on similar tasks may stem from uncontrolled comparisons.

The contributions of this work are as follows:
\begin{itemize}
    \item A \textbf{parameter matched} comparison of quantum and classical latent generators that isolates the quantum contribution to medical image augmentation, rather than confounding it with model
    capacity.
    \item A \textbf{label efficiency study} that sweeps labeled data fractions from 5\% to 100\% with eight seeds and paired significance tests, identifying where augmentation helps.
    \item A \textbf{joint evaluation} of generation quality (intra-set diversity and latent distribution overlap) and downstream classification, linking sample quality to utility.
    \item A \textbf{benchmark} and the empirical finding that, under control, quantum and classical latent GAN
    augmentation performs equivalently and act as regularization in the low-data regime.
\end{itemize}

\section{Related Work}

\subsection{Generative Augmentation in Medical Imaging}
GAN-based augmentation has been applied across medical modalities to expand limited datasets and rebalance under-represented classes \cite{sun2022hierarchical,fan2023u}. Brain-MRI classification in particular has benefited from synthetic-image pipelines, with several works reporting
accuracy gains from GAN- or diffusion-based augmentation
\cite{zhong2026image,muller2023multimodal}. A recurring design encodes images into a compact latent space and models that space generatively, which stabilizes adversarial training and reduces the cost of high resolution synthesis. The same latent space strategy underlies our pipeline and, importantly, makes a quantum generator practical: a shallow circuit need only produce a low-dimensional code rather than a full image.

\subsection{Quantum Machine Learning and Generative Models}
Variational quantum architectures have emerged as a foundational paradigm near term quantum machine learning by offering trainable quantum circuits that combine data encoding, parameterized rotations, entangling layers, and classical optimization. 
Quantum GANs instantiate this idea for generation, replacing part of the generator with such a circuit
\cite{ngo2023survey}. Their appeal is two fold entangled states can represent correlations compactly, and the parameter count is small but their trainability is limited at scale by phenomena such as barren plateaus \cite{larocca2025barren}, and shallow circuits on a handful of qubits have restricted expressivity. Subsequent work has scaled quantum and quantum inspired generators toward higher resolution \cite{khatun2025quantum} and recent medical imaging applications combine such circuits with classical decoders and report improvements over classical baselines \cite{hanafi2025quantum, nandal2025image}.
\subsection{The Evaluation Gap}
The comparisons in prior quantum augmentation studies are frequently drawn from single runs, do not equalize the parameter budgets of the quantum and classical generators and do not report where in the data regime the benefit arises. Generation quality is rarely characterized, so it is
unclear whether reported gains reflect realistic synthesis or incidental regularization. Our work differs by comparing the quantum based models with classical models as a controlled experiment: matched parameter counts, repeated
seeds, a data fraction sweep, joint generation quality and downstream evaluation.

\section{Background}

\subsection{Variational Autoencoder (VAE)}
A VAE \cite{jabbar2025fusion} learns an encoder $q_\phi(z\mid x)=\mathcal{N}\!\big(\mu_\phi(x),\,\mathrm{diag}\,\sigma_\phi^2(x)\big)$
and a decoder $p_\psi(x\mid z)$, regularizing the latent space toward a
standard normal prior $p(z)=\mathcal{N}(0,I)$. With a reconstruction term
and a Kullback-Leibler (KL) penalty, the training objective is
\begin{equation}
\mathcal{L}_{\mathrm{VAE}} = \lVert x-\hat{x}\rVert_2^2
\;+\;\beta\, D_{\mathrm{KL}}\!\big(q_\phi(z\mid x)\,\Vert\,p(z)\big),
\end{equation}
where the KL term has the closed form
$-\tfrac{1}{2}\sum_{j=1}^{d}\big(1+\log\sigma_j^2-\mu_j^2-\sigma_j^2\big)$.
We use the posterior mean $\mu_\phi(x)$ as the latent code. The penalty $\beta$ controls how strongly the aggregated posterior is pulled towards the prior, which determines whether sampled or generated codes decode to
coherent images.

\subsection{Conditional WGAN-GP}
A Wasserstein GAN \cite{peketi2023flwgan} replaces the Jensen Shannon objective
with the Earth Mover distance, estimated by a 1-Lipschitz critic $D_w$.
Enforcing the Lipschitz constraint via a gradient penalty \cite{roy2023novel} and conditioning on the class label $y$, the critic and generator objectives are
\begin{align}
\mathcal{L}_D &= \mathbb{E}_{\tilde{z}}\big[D_w(\tilde{z},y)\big]
 - \mathbb{E}_{z}\big[D_w(z,y)\big]
 + \lambda\,\mathbb{E}_{\hat{z}}\!\big[(\lVert\nabla_{\hat{z}}D_w(\hat{z},y)\rVert_2-1)^2\big],\\
\mathcal{L}_G &= -\,\mathbb{E}_{\tilde{z}}\big[D_w(\tilde{z},y)\big],
\end{align}
where $z$ are real latent codes, $\tilde{z}=G_\theta(\cdot,y)$ are
generated codes, and $\hat{z}=\epsilon z+(1-\epsilon)\tilde{z}$ with
$\epsilon\sim U(0,1)$. We use $\lambda=10$ and five critic updates per generator update.

\subsection{Variational Quantum Generator}
On $n$ qubits, an input vector $a\in\mathbb{R}^{n}$ is encoded by single-qubit rotations $S(a)=\bigotimes_{i=1}^{n}R_y(a_i)$, followed by a
depth-$L$ trainable block $U(\theta)$ of $R_y/R_z$ rotations and a ring of CNOT entanglers. The prepared state is
\begin{equation}
\lvert\psi(a,\theta)\rangle = U(\theta)\,S(a)\,\lvert 0\rangle^{\otimes n},
\end{equation}
and the circuit outputs the Pauli-$Z$ expectations
$o_i=\langle\psi\lvert Z_i\rvert\psi\rangle\in[-1,1]$. A small classical
head $h$ maps these measurements, concatenated with a label embedding
$e_y$, to the latent code: $\tilde{z}=h([o_1,\dots,o_n,e_y])$. In the
classical generator the measurements $o$ are replaced by Gaussian noise of
the same dimension, with $h$ and $e_y$ unchanged.

\section{Methodology}

\subsection{Overview}
The pipeline has three stages: (i) a VAE learns a low-dimensional latent
representation of brain-MRI images; (ii) a conditional WGAN-GP is trained
in that latent space using either the quantum or the classical generator
of matched size; (iii) synthetic latents are decoded into images and used
to augment a downstream classifier. The quantum and classical arms are
identical except for the generator, isolating the quantum contribution.
Algorithm~\ref{alg:protocol} summarizes the controlled protocol.

\begin{algorithm}[t]
\caption{Controlled label-efficiency benchmark}
\label{alg:protocol}
\begin{algorithmic}[1]
\STATE \textbf{Input:} dataset $\mathcal{D}$, fractions $\mathcal{F}$, seeds $\mathcal{S}$
\FOR{each fraction $f\in\mathcal{F}$}
  \STATE $\mathcal{D}_f\leftarrow$ class-stratified subsample of $\mathcal{D}_{\mathrm{train}}$
  \STATE train VAE on $\mathcal{D}_f$; encode latents $Z_f=\mu_\phi(\mathcal{D}_f)$
  \STATE train $G_{\mathrm{c}}$ and $G_{\mathrm{q}}$ (WGAN-GP) on $Z_f$
  \STATE decode synthetic sets $\mathcal{X}_{\mathrm{c}},\mathcal{X}_{\mathrm{q}}$
  \STATE record diversity and latent-overlap metrics
  \FOR{each seed $s\in\mathcal{S}$}
    \STATE train + evaluate classifier on $\mathcal{D}_f$, $\mathcal{D}_f\!\cup\!\mathcal{X}_{\mathrm{c}}$, $\mathcal{D}_f\!\cup\!\mathcal{X}_{\mathrm{q}}$
  \ENDFOR
\ENDFOR
\STATE \textbf{Output:} accuracy/F1 (mean$\pm$std) and corrected paired tests
\end{algorithmic}
\end{algorithm}
\begin{figure}[t]
\centering
\resizebox{0.8\linewidth}{!}{%
\begin{tikzpicture}[
  font=\small,>={Stealth[length=2.4mm]},
  data/.style={draw,thick,rounded corners=1pt,align=center,fill=gray!8,minimum height=11mm,minimum width=24mm,inner sep=3pt},
  block/.style={draw,thick,rounded corners=3pt,align=center,fill=blue!6,minimum height=11mm,minimum width=24mm,inner sep=3pt},
  proc/.style={draw,thick,rounded corners=3pt,align=center,fill=green!8,minimum height=11mm,minimum width=24mm,inner sep=3pt},
  gen/.style={draw,thick,rounded corners=3pt,align=center,minimum width=34mm,minimum height=8mm,inner sep=2pt},
  arr/.style={->,thick},
]
\node[data] (img) {Real MRI\\$x$};
\node[block,right=12mm of img] (enc) {VAE encoder\\$q_\phi$};
\node[gen,fill=orange!18,right=18mm of enc] (gq) {Quantum gen.\ $G_q$};
\node[gen,fill=blue!12,below=2.4mm of gq] (gc) {Classical gen.\ $G_c$};
\node[gen,fill=gray!14,below=2.4mm of gc] (crit) {Critic $D_w$};
\begin{scope}[on background layer]
  \node[draw,thick,rounded corners=4pt,fill=orange!6,fit=(gq)(gc)(crit),inner sep=5mm] (gan) {};
\end{scope}
\node[anchor=south,font=\bfseries] at ([yshift=0.8mm]gan.north) {Conditional WGAN-GP};
\node[block,right=18mm of gan] (dec) {VAE decoder\\$p_\psi$};
\node[data,right=12mm of dec] (syn) {Synthetic\\MRI};
\draw[arr] (img) -- (enc);
\draw[arr] (enc) -- node[above,font=\footnotesize]{$z\in\mathbb{R}^{16}$} (gan.west);
\draw[arr] (gan.east) -- node[above,font=\footnotesize]{$\tilde z$} (dec);
\draw[arr] (dec) -- (syn);
\draw[arr,gray] (gq.east) to[out=0,in=60] (crit.east);
\draw[arr,gray] (gc.east) to[out=0,in=30] (crit.east);
\node[font=\footnotesize] (lbl) at ([yshift=-8mm]gan.south) {class label $y$};
\draw[arr] (lbl) -- (gan.south);
\node[font=\bfseries,anchor=west] at ([yshift=4mm]img.north west) {(a) Methodology pipeline};
\node[proc,below=34mm of syn] (aug) {Augment\\Real $\cup$ Synthetic};
\node[block,left=14mm of aug] (clf) {ResNet-18\\classifier};
\node[proc,left=14mm of clf,minimum width=44mm] (eval) {Accuracy / F1\\Diversity (SSIM, pixel-std)\\t-SNE overlap};
\draw[arr] (syn.south) -- (aug.north);
\draw[arr] (aug) -- (clf);
\draw[arr] (clf) -- (eval);

\coordinate (bTL) at ($(img.west |- eval.south)+(0,-26mm)$);
\begin{scope}[shift={(bTL)},x=2.25cm,y=1.05cm]
  \def\wireL{7.6}
  \node[font=\bfseries,anchor=west] at (-0.05,1.85) {(b) Variational quantum generator $G_q$};
  \foreach \i in {0,1,2,3}{ \draw[thick] (0,-\i) -- (\wireL,-\i); \node[left] at (0,-\i) {$|0\rangle$}; }
  \foreach \i in {0,1,2,3}{ \node[draw,thick,fill=orange!12,minimum size=10mm] at (1.15,-\i) {$R_y(a_{\the\numexpr\i+1})$}; }
  \foreach \i in {0,1,2,3}{ \node[draw,thick,fill=blue!10,minimum size=10mm] at (2.65,-\i) {$R_yR_z$}; }
  \foreach \a/\b/\x in {0/1/3.7, 1/2/4.15, 2/3/4.6}{
    \fill (\x,-\a) circle (3pt); \draw[thick] (\x,-\a) -- (\x,-\b);
    \draw[thick] (\x,-\b) circle (5pt); \draw[thick] (\x,-\b-5pt) -- (\x,-\b+5pt); }
  \fill (5.1,-3) circle (3pt); \draw[thick] (5.1,-3) -- (5.1,0);
  \draw[thick] (5.1,0) circle (5pt); \draw[thick] (5.1,-5pt) -- (5.1,5pt);
  \foreach \i in {0,1,2,3}{ \node[draw,thick,fill=gray!12,minimum width=12mm,minimum height=9mm] (m\i) at (6.0,-\i) {$\langle Z_{\the\numexpr\i+1}\rangle$}; }
  \node[font=\footnotesize] at (1.15,0.85) {encode $S(a)$};
  \node[font=\footnotesize] at (3.9,0.85) {trainable $U(\theta)$\ ($\times L{=}2$)};
  \node[font=\footnotesize] at (6.0,0.85) {measure};
  \node[draw,thick,rounded corners=3pt,fill=green!8,align=center,right=9mm of m0] (head) {classical head\\$h\ \to\ \tilde z$};
  \draw[arr] (m0.east) -- (head.west);
\end{scope}
\end{tikzpicture}
}
\caption{Overview of the controlled benchmark. (a) Brain-MRI images are encoded by a VAE into a 16-D latent space; a conditional WGAN-GP trains \emph{either} a quantum or a parameter-matched classical generator against a shared critic; decoded synthetic images augment the real data for a ResNet-18 classifier, evaluated by accuracy/F1, intra-set diversity, and latent-distribution overlap. (b) The variational quantum generator $G_q$: $R_y$ angle encoding $S(a)$, a depth-$L{=}2$ block of $R_y/R_z$ rotations with a CNOT entangling ring $U(\theta)$, Pauli-$Z$ measurement, and a classical head mapping the four expectations to the latent code $\tilde z$.}
\label{fig:method}
\end{figure}
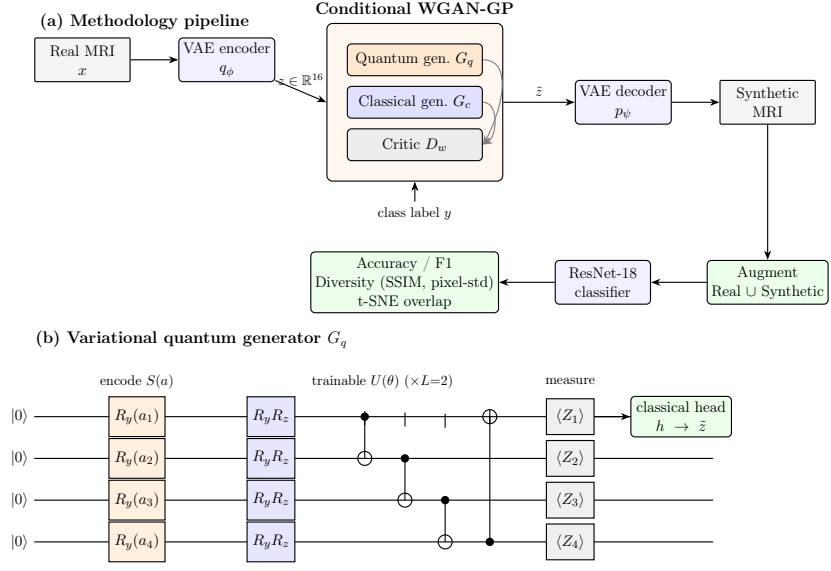

\subsection{Latent Representation}
Grayscale images are resized to $128\times128$ and normalized to
$[-1,1]$. The convolutional VAE encodes each image to a $16$-dimensional
latent code through four strided convolutional blocks
($1\!\to\!32\!\to\!64\!\to\!128\!\to\!256$ channels); the decoder mirrors
the encoder with transposed convolutions and a $\tanh$ output. We set
$\beta=10^{-3}$, which we found sufficient to keep decoded codes coherent
without over-smoothing reconstructions.

\subsection{Generators and Critic}
The quantum generator uses $n=4$ qubits and depth $L=2$
\cite{shapiro2025hybrid}; the classical generator matches its conditioning and
output head. The critic is a LayerNorm MLP taking a latent vector concatenated with a class embedding. The two generators contain $1648$ and
$1632$ trainable parameters respectively (Table~\ref{tab:params}), so any
performance difference cannot be attributed to capacity.

\subsection{Synthetic Augmentation and Classification}
For each class, $300$ latent codes are generated and decoded into images.
The downstream classifier is a pretrained ResNet-18 \cite{xu2023resnet} with grayscale input and a four-class head. We compare three training sets:
real-only, real $+$ classical-synthetic, and real $+$ quantum-synthetic,
holding all other settings fixed.

\subsection{Generation Quality Metrics}
Rather than ImageNet-based Fr\'echet Inception Distance, which is poorly calibrated for grayscale medical images \cite{woodland2024feature}, we characterize generation quality with (i) intra-set diversity, the mean pairwise SSIM among synthetic samples and the per pixel standard deviation across the synthetic batch; and (ii) latent distribution overlap, visualized by projecting real and synthetic latents with t-SNE \cite{cai2022theoretical}. High pairwise SSIM or low pixel standard deviation indicates mode collapse.

\section{Experimental Setup}
\label{sec:setup}

\subsection{Dataset and Preprocessing}
We use the publicly available Brain Tumor MRI Dataset
\cite{hira2025brain}, which consists of $12{,}064$ T1 weighted
contrast enhanced brain MRI images divided into four classes (glioma, meningioma,
no-tumor, pituitary). We utilize the dataset's fixed
partition of $9{,}650$ training and $2{,}414$ test images; the test split
(glioma~755, meningioma~546, no-tumor~487, pituitary~626) is held fixed
across all conditions and is never used for training or model selection.
Images vary in size and are resized to $128\times128$, converted to
grayscale, and normalized to $[-1,1]$. No separate validation set is used:
each model is trained for a fixed budget (Section~\ref{sec:impl}) and the
held out test split is evaluated exactly once per run, so no
test set based model selection occurs. Class stratified sub-sampling for the
low data fractions is applied to the \emph{training} split only; the test
split is identical in every experiment, precluding leakage across
conditions.

\subsection{Statistical Analysis}
For each data fraction and pair of conditions we form eight seed matched accuracy observations (one per random seed) and apply a two sided paired $t$-test. Because $n=8$ is small, we additionally report the non-parametric
Wilcoxon signed rank test and verify approximate normality of the paired differences with the Shapiro Wilk test. To control for the multiple comparisons across fractions and condition pairs, we apply Holm Bonferroni correction. No comparison survives correction: the smallest corrected
$p$-value is $0.62$ (classical vs.\ real at $25\%$), and the near threshold quantum vs classical case at $25\%$ rises to $0.68$; all others reach $1.0$.
The Wilcoxon tests agree with the $t$-tests throughout (quantum vs.\ classical at $25\%$, $p=0.078$).

\subsection{Implementation and Training}
\label{sec:impl}
Quantum circuits are simulated with a state-vector backend
\cite{chen2024quantum}; all models are implemented in PyTorch and trained on
a single GPU. The VAE is trained for $12$ epochs (AdamW, lr $10^{-3}$);
each WGAN-GP for $80$ epochs (Adam, lr $10^{-4}$, $\beta=(0.5,0.9)$, batch
$64$); each classifier for $6$ epochs (AdamW, lr $10^{-4}$). For the main
results, generators are trained once per fraction with a fixed seed and the
downstream classifier is repeated over eight seeds; we additionally repeat
generator training over three seeds in the low data regime to quantify
generator initialization variance (Section~\ref{sec:genvar}).
Concretely, the VAE and both WGAN-GP generators are trained exactly once per
data fraction (fixed seed), and the synthetic image sets are generated once
and reused unchanged across the eight classifier runs. The only source of
variation in Tables~\ref{tab:acc} and~\ref{tab:f1} is therefore the
classifier's random seed (the initialization of its new head and the training
data ordering); no upstream component is retrained across those seeds. The
three-seed generator sweep of Section~\ref{sec:genvar} is the sole experiment
in which the generators are retrained.

\begin{table}[t]
\centering
\caption{Trainable parameter counts. Generators are matched.}
\label{tab:params}
\begin{tabular}{lc}
\hline
Component & Parameters \\
\hline
Quantum generator   & 1{,}648 \\
Classical generator & 1{,}632 \\
Critic              & 12{,}993 \\
\hline
\end{tabular}
\end{table}

\section{Results and Discussion}

\subsection{Downstream Classification}
\begin{figure}[!htb]
\centering
\includegraphics[width=\textwidth]{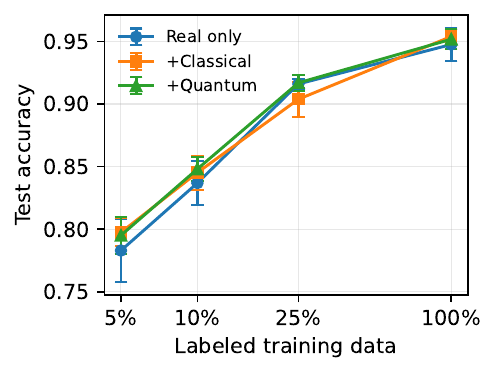}
\caption{Test accuracy versus available labeled data fraction for the three
conditions (mean $\pm$ std, 8 seeds). The conditions overlap within
variance at every fraction, and augmentation provides no significant gain.}
\label{fig:acc}
\end{figure}

Table~\ref{tab:acc} and Fig.~\ref{fig:acc} report test accuracy across
data fractions, and Table~\ref{tab:f1} reports the corresponding weighted
F1. No augmentation variant significantly outperforms real-data-only
training at any fraction: all paired $t$-tests of quantum versus real-only
yield $p>0.05$, and as Fig.~\ref{fig:acc} makes visually clear, the three
conditions overlap within one standard deviation at every operating point.
The quantum and classical generators are likewise statistically
indistinguishable. The only near-significant uncorrected difference appears
at $25\%$ ($p=0.062$, quantum versus classical), where the classical
generator degraded accuracy below the real baseline while the quantum
generator didn't had any observation about robustness rather than a
performance gain and this difference does not survive multiple comparison
correction (Section~\ref{sec:setup}). Augmentation effects are largest,
though still not significant, in the low data regime and vanish as real data
becomes abundant, consistent with augmentation acting as a regularizer
rather than a source of new information. The F1 scores in
Table~\ref{tab:f1} track accuracy closely, indicating the conclusions are
not an artifact of class imbalance.

\begin{table}[t]
\centering
\caption{Test accuracy (mean $\pm$ std over 8 seeds). $p$ values are
uncorrected paired $t$-tests; none are significant at $\alpha=0.05$ except
the near threshold quantum vs classical case at $25\%$, which does not
survive Holm Bonferroni correction.}
\label{tab:acc}
\setlength{\tabcolsep}{4pt}
\begin{tabular}{lccccc}
\hline
Frac. & Real & +Classical & +Quantum & $p_{Q,R}$ & $p_{Q,C}$ \\
\hline
5\%   & .783$\pm$.025 & .798$\pm$.011 & .795$\pm$.015 & .305 & .740 \\
10\%  & .837$\pm$.018 & .845$\pm$.014 & .848$\pm$.010 & .173 & .525 \\
25\%  & .916$\pm$.004 & .904$\pm$.014 & .917$\pm$.006 & .762 & .062 \\
100\% & .948$\pm$.013 & .954$\pm$.006 & .952$\pm$.008 & .473 & .530 \\
\hline
\end{tabular}
\end{table}

\begin{table}[t]
\centering
\caption{Weighted F1 (mean $\pm$ std over 8 seeds). Trends mirror accuracy.}
\label{tab:f1}
\setlength{\tabcolsep}{6pt}
\begin{tabular}{lccc}
\hline
Frac. & Real & +Classical & +Quantum \\
\hline
5\%   & .784$\pm$.024 & .793$\pm$.011 & .792$\pm$.014 \\
10\%  & .835$\pm$.019 & .844$\pm$.013 & .848$\pm$.009 \\
25\%  & .915$\pm$.004 & .903$\pm$.014 & .916$\pm$.007 \\
100\% & .947$\pm$.014 & .954$\pm$.006 & .951$\pm$.008 \\
\hline
\end{tabular}
\end{table}

\subsection{Generator-Initialization Variance}
\label{sec:genvar}
Because GAN training can be unstable, we repeat generator training over three
random seeds in the low data regime ($5\%$ and $10\%$ of the data), the
configuration in which adversarial training is least stable and
initialization variance is largest with the classifier seed held fixed.
Across generator seeds, downstream accuracy varies by at most $0.023$ at
$5\%$ and $0.010$ at $10\%$, comparable to or below the classifier-seed
variability reported in Table~\ref{tab:acc}. No significance conclusion
changes, so the parity outcome is not an artifact of a single generator
initialization; since the higher fractions train more stably, their
initialization variance can only be smaller.

\subsection{Generation Quality}
\begin{figure}[!ht]
\centering
\includegraphics[width=\textwidth]{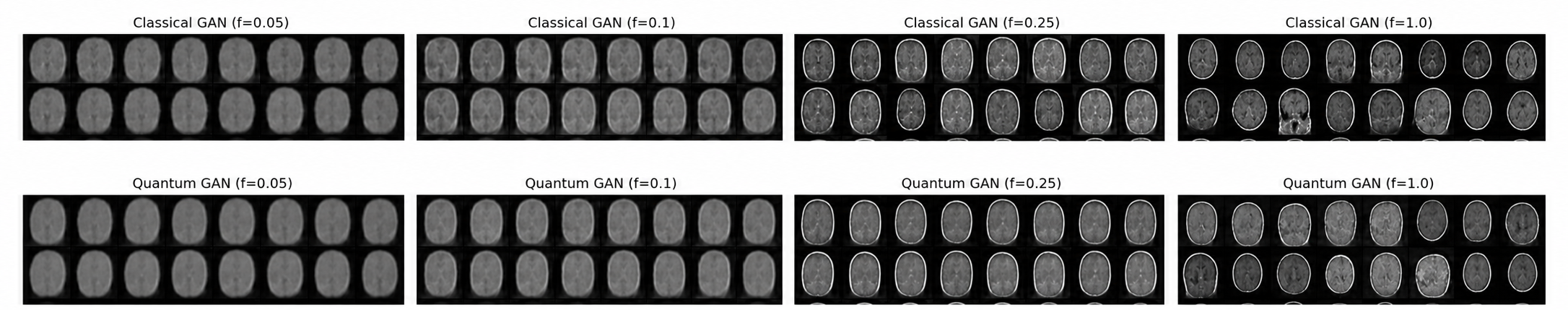}
\caption{Decoded synthetic samples (top: classical generator; bottom:
quantum generator) at $5\%$, $10\%$, $25\%$, and $100\%$ of the training
data. Both generators collapse to near identical, low contrast images at
low data and recover structure and diversity only as data increases; the
quantum generator is marginally more collapsed at the smallest fractions.}
\label{fig:samples}
\end{figure}

Table~\ref{tab:div} and Fig.~\ref{fig:samples} show that synthetic
diversity scales with available data: mean pairwise SSIM falls from
$\sim0.95$--$0.98$ at $5\%$ (nearly identical samples) to
$\sim0.46$--$0.48$ at full data, while per pixel standard deviation rises
correspondingly. The sample grids in Fig.~\ref{fig:samples} make this
concrete: at $5\%$ and $10\%$ both generators emit near identical, blurry
images and only at $25\%$ and $100\%$ do skull
boundaries, internal texture and pose variation emerge. The quantum
generator is consistently \emph{less} diverse than the classical one
(pairwise SSIM $0.982$ vs.\ $0.953$ at $5\%$), yet achieves
comparable downstream utility, further evidence that the augmentation
benefit is not driven by sample fidelity.

\begin{table}[!htb]
\centering
\caption{Intra-set diversity. Higher pairwise SSIM or lower pixel-std
indicates stronger mode collapse.}
\label{tab:div}
\setlength{\tabcolsep}{4pt}
\begin{tabular}{lcccc}
\hline
& \multicolumn{2}{c}{Pairwise SSIM $\downarrow$} & \multicolumn{2}{c}{Pixel std $\uparrow$} \\
Frac. & Classical & Quantum & Classical & Quantum \\
\hline
5\%   & 0.953 & 0.982 & 0.025 & 0.015 \\
10\%  & 0.894 & 0.945 & 0.033 & 0.022 \\
25\%  & 0.638 & 0.713 & 0.052 & 0.039 \\
100\% & 0.463 & 0.479 & 0.098 & 0.096 \\
\hline
\end{tabular}
\end{table}

\subsection{Latent-Distribution Analysis}
\begin{figure}[!htb]
\centering
\includegraphics[width=\textwidth]{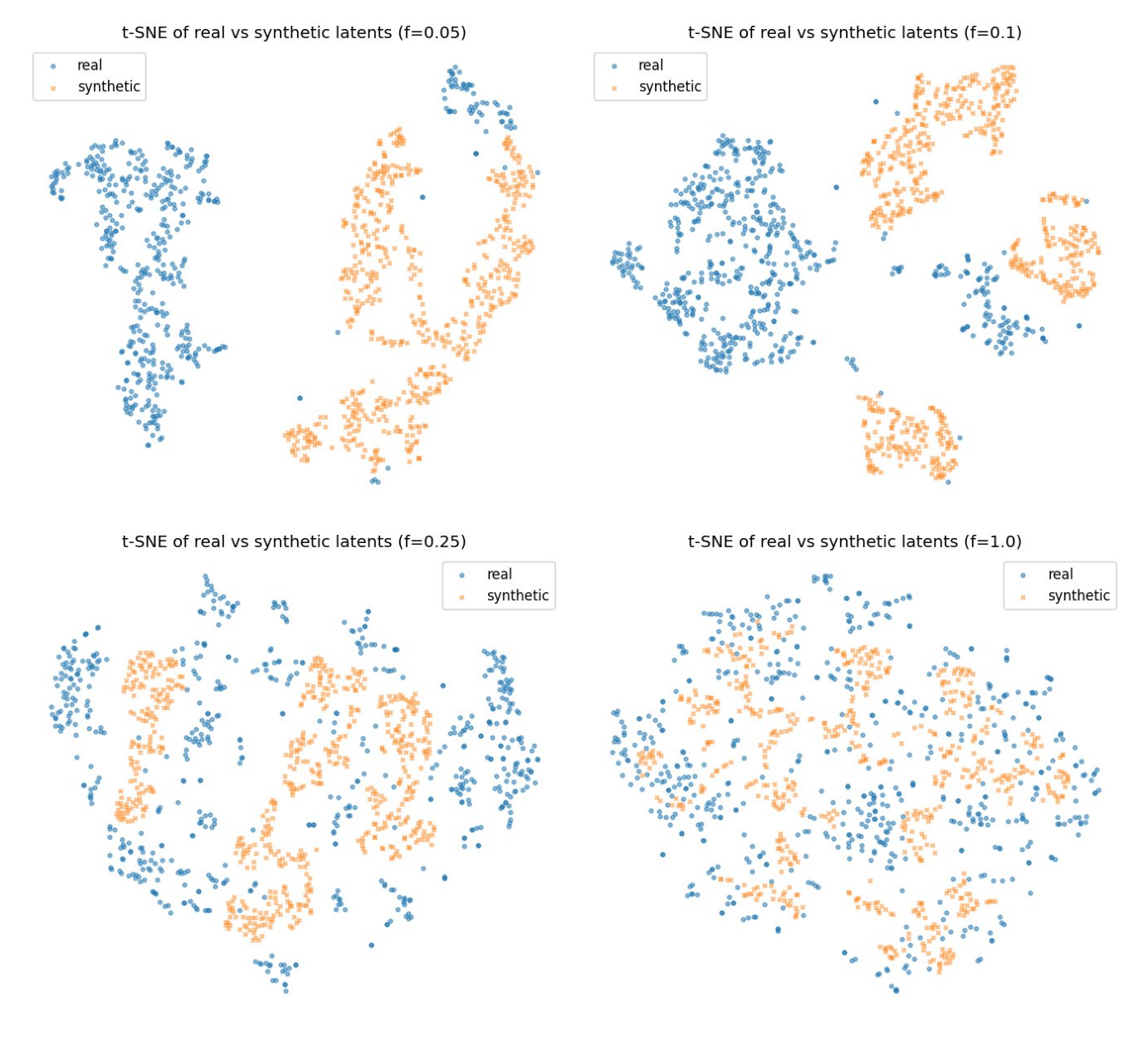}
\caption{t-SNE of real (blue) and synthetic (orange) latents for the
quantum generator at $5\%$, $10\%$, $25\%$, and $100\%$ data. Real--synthetic
overlap improves monotonically with available data; the classical
generator behaves almost identically.}
\label{fig:tsne}
\end{figure}

Fig.~\ref{fig:tsne} projects real and synthetic latents at each fraction. At
$5\%$ and $10\%$ the synthetic codes (orange) form regions largely disjoint
from the real data (blue), indicating that generated samples are
off-distribution exactly where labeled data is scarce. The overlap improves
monotonically as data grows, becoming substantial at $100\%$. This explains
the downstream pattern: in the low-data regime the generator cannot match the
real distribution, so synthetic samples can only regularize the decision
boundary; in the high-data regime the distribution is matched but
augmentation is no longer needed.

\subsection{Why Quantum and Classical Reach Parity}
Three factors plausibly explain the absence of a quantum effect at this
scale. First, the quantum circuit contributes only four bounded
expectation values, which a classical fully connected head reshapes into
the latent code; the head, common to both arms, carries most of the
modeling burden, so the four quantum outputs are easily matched by four
Gaussian noise dimensions. Second, a depth-$2$, four-qubit circuit has
limited expressivity and weak effective use of entanglement, and deeper
circuits face trainability barriers such as barren plateaus
\cite{larocca2025barren}. Third, operating in a learned latent space decouples
the generator from pixel-level structure, so any quantum inductive bias is
diluted by the decoder. Together these mean the quantum component behaves,
empirically, as an alternative low-dimensional noise source rather than a
source of richer correlations.

\subsection{Implications for Evaluating Quantum Augmentation}
Our results show how easily an uncontrolled protocol could have produced an
apparent quantum advantage: a single favorable seed at $10\%$, or a
comparison against the classical generator at $25\%$ (where it degraded),
would each suggest the quantum model wins. Only the matched parameter,
multi seed, regime spanning design with corrected significance testing
reveals these as noise or as classical fragility rather than quantum
strength. We therefore recommend that future quantum augmentation studies
report parameter matched baselines, seed averaged results with significance
tests, and a data regime sweep and that they characterize generation
quality directly rather than inferring it from downstream accuracy alone.

\subsection{Reproducibility}
All hyperparameters are listed in Section~\ref{sec:impl} and
Table~\ref{tab:params}; the test split is fixed; generators use fixed and
swept seeds and classifiers eight seeds; and the full protocol
(Algorithm~\ref{alg:protocol}) and code are released (see Data and Code
Availability).

\section{Limitations}
Our study uses an ideal state vector simulator, a shallow four qubit
circuit, and a single dataset; deeper circuits, alternative
encodings and additional modalities may have different results. Operating in a
learned latent space decouples the quantum model from pixel level fidelity;
pixel space quantum generation may behave differently. Because the circuits
are evaluated on a noiseless simulator, the reported quantum results
represent a best case: on NISQ hardware, gate noise, decoherence, and
finite shot estimation of the Pauli-$Z$ expectations would inject additional
stochasticity and reduce the effective expressivity of the generator. Such
effects would be expected to widen, rather than close, the gap relative to
the classical baseline, so the parity we observe is unlikely to be masking a
latent hardware advantage; confirming this on real devices is left to future
work.

\section{Conclusion}
We presented a controlled, parameter matched, multi seed benchmark of
quantum versus classical latent GAN augmentation for brain MRI
classification across a range of labeled data fractions. Under these
controls, quantum and classical augmentation perform equivalently and
neither significantly improves over real data only training; any low data
benefit is consistent with regularization rather than realistic synthesis,
as evidenced by off-distribution, mode collapsed samples precisely where
augmentation would be expected to help. We emphasize that this negative
result is specific to the tested configuration; a four-qubit, depth-$2$
quantum generator operating in a learned latent space on a single brain-MRI
dataset and should not be read as a general claim about quantum generative
augmentation. We hope the protocol serves as a
testbed for evaluation of quantum generative models in medical
imaging. Future work includes larger circuits, real quantum hardware,
domain appropriate fidelity metrics and extension to additional imaging
modalities.

\section*{Data and Code Availability}
The dataset is publicly available \cite{hira2025brain}. The full pipeline, configuration files and the protocol of Algorithm~\ref{alg:protocol} are hosted in a public repository at \url{https://github.com/iMujtabaSyed/Quantum-Latent-GAN-Augmentation-for-Brain-MRI}.

\bibliographystyle{ieeetr}
\bibliography{bib}

@inproceedings{al2023brain,
  title={Brain tumor recognition from MRI using deep learning with data balancing methods and its explainability with AI},
  author={Al Noman, Abdullah and Arif, Abu Shamim Mohammad},
  booktitle={International Conference on Image Processing and Capsule Networks},
  pages={523--538},
  year={2023},
  organization={Springer}
}

@article{hira2025brain,
  title={Brain tumor mri dataset (glioma meningioma pituitary no tumor)},
  author={Hira, MI Kabir and Hossain, MS and Bithee, MM Akter and Sara, U and Hasan, MM and Towsif, AA and Ahmed, MK},
  journal={Mendeley Data},
  volume={1},
  pages={2025},
  year={2025}
}

@article{guo2023medgan,
  title={MedGAN: An adaptive GAN approach for medical image generation},
  author={Guo, Kehua and Chen, Jie and Qiu, Tian and Guo, Shaojun and Luo, Tao and Chen, Tianyu and Ren, Sheng},
  journal={Computers in Biology and Medicine},
  volume={163},
  pages={107119},
  year={2023},
  publisher={Elsevier}
}

@article{makhlouf2023use,
  title={The use of generative adversarial networks in medical image augmentation},
  author={Makhlouf, Ahmed and Maayah, Marina and Abughanam, Nada and Catal, Cagatay},
  journal={Neural Computing and Applications},
  volume={35},
  number={34},
  pages={24055--24068},
  year={2023},
  publisher={Springer}
}

@article{chang2024latent,
  title={Latent style-based quantum gan for high-quality image generation},
  author={Chang, Su Yeon and Thanasilp, Supanut and Saux, Bertrand Le and Vallecorsa, Sofia and Grossi, Michele},
  journal={arXiv preprint arXiv:2406.02668},
  year={2024}
}

@article{tsang2023hybrid,
  title={Hybrid quantum--classical generative adversarial network for high-resolution image generation},
  author={Tsang, Shu Lok and West, Maxwell T and Erfani, Sarah M and Usman, Muhammad},
  journal={IEEE Transactions on Quantum Engineering},
  volume={4},
  pages={1--19},
  year={2023},
  publisher={IEEE}
}

@article{sun2022hierarchical,
  title={Hierarchical amortized GAN for 3D high resolution medical image synthesis},
  author={Sun, Li and Chen, Junxiang and Xu, Yanwu and Gong, Mingming and Yu, Ke and Batmanghelich, Kayhan},
  journal={IEEE journal of biomedical and health informatics},
  volume={26},
  number={8},
  pages={3966--3975},
  year={2022},
  publisher={IEEE}
}

@article{fan2023u,
  title={U-Patch GAN: A medical image fusion method based on GAN},
  author={Fan, Chao and Lin, Hao and Qiu, Yingying},
  journal={Journal of Digital Imaging},
  volume={36},
  number={1},
  pages={339--355},
  year={2023},
  publisher={Springer}
}

@article{muller2023multimodal,
  title={A multimodal comparison of latent denoising diffusion probabilistic models and generative adversarial networks for medical image synthesis},
  author={M{\"u}ller-Franzes, Gustav and Niehues, Jan Moritz and Khader, Firas and Arasteh, Soroosh Tayebi and Haarburger, Christoph and Kuhl, Christiane and Wang, Tianci and Han, Tianyu and Nolte, Teresa and Nebelung, Sven and others},
  journal={Scientific reports},
  volume={13},
  number={1},
  pages={12098},
  year={2023},
  publisher={Nature Publishing Group UK London}
}

@article{zhong2026image,
  title={Image enhancement for accelerated MRI using a joint GAN and diffusion model framework},
  author={Zhong, Quan and Zhu, Shipai and He, Jinrong and Wang, Haitao and Zhong, Renming},
  journal={Medical Physics},
  volume={53},
  number={1},
  pages={e70242},
  year={2026},
  publisher={Wiley Online Library}
}

@article{jabbar2025fusion,
  title={Fusion-aware quantum variational autoencoder for brain-heart signal modeling in mental health applications},
  author={Jabbar, Ayesha and Jianjun, Huang and Jabbar, Muhammad Kashif and Mahmood, Tariq and Haider, Syed Mujtaba},
  journal={Journal of King Saud University Computer and Information Sciences},
  volume={37},
  number={9},
  pages={1--21},
  year={2025},
  publisher={Springer}
}

@article{ngo2023survey,
  title={A survey of recent advances in quantum generative adversarial networks},
  author={Ngo, Tuan A and Nguyen, Tuyen and Thang, Truong Cong},
  journal={Electronics},
  volume={12},
  number={4},
  pages={856},
  year={2023},
  publisher={MDPI}
}

@article{larocca2025barren,
  title={Barren plateaus in variational quantum computing},
  author={Larocca, Martin and Thanasilp, Supanut and Wang, Samson and Sharma, Kunal and Biamonte, Jacob and Coles, Patrick J and Cincio, Lukasz and McClean, Jarrod R and Holmes, Zo{\"e} and Cerezo, Marco},
  journal={Nature Reviews Physics},
  volume={7},
  number={4},
  pages={174--189},
  year={2025},
  publisher={Nature Publishing Group UK London}
}

@article{khatun2025quantum,
  title={Quantum generative learning for high-resolution medical image generation},
  author={Khatun, Amena and Yeter Aydeniz, K{\"u}bra and Weinstein, Yaakov S and Usman, Muhammad},
  journal={Machine Learning: Science and Technology},
  volume={6},
  number={2},
  pages={025032},
  year={2025},
  publisher={IOP Publishing}
}

@inproceedings{hanafi2025quantum,
  title={A Quantum-Classical GAN Approach for HighFidelity Brain MRI Resolution Enhancement},
  author={Hanafi, Basil and Ali, Mohammad},
  booktitle={2025 12th International Conference on Computing for Sustainable Global Development (INDIACom)},
  pages={1--7},
  year={2025},
  organization={IEEE}
}

@article{nandal2025image,
  title={Image denoising using quantum deep convolutional generative adversarial network for medical images},
  author={Nandal, Priyanka and Pahal, Sudesh and Upadhyay, Govind Murari},
  journal={International Journal of Computational Intelligence Systems},
  volume={18},
  number={1},
  pages={190},
  year={2025},
  publisher={Springer}
}

@inproceedings{peketi2023flwgan,
  title={Flwgan: Federated learning with wasserstein generative adversarial network for brain tumor segmentation},
  author={Peketi, Divya and Chalavadi, Vishnu and Mohan, C Krishna and Chen, Yen Wei},
  booktitle={2023 International Joint Conference on Neural Networks (IJCNN)},
  pages={1--8},
  year={2023},
  organization={IEEE}
}

@article{roy2023novel,
  title={A novel conditional Wasserstein deep convolutional generative adversarial network},
  author={Roy, Arunava and Dasgupta, Dipankar},
  journal={IEEE Transactions on Artificial Intelligence},
  year={2023},
  publisher={IEEE}
}

@inproceedings{woodland2024feature,
  title={Feature extraction for generative medical imaging evaluation: New evidence against an evolving trend},
  author={Woodland, McKell and Castelo, Austin and Al Taie, Mais and Albuquerque Marques Silva, Jessica and Eltaher, Mohamed and Mohn, Frank and Shieh, Alexander and Kundu, Suprateek and Yung, Joshua P and Patel, Ankit B and others},
  booktitle={International Conference on Medical Image Computing and Computer-Assisted Intervention},
  pages={87--97},
  year={2024},
  organization={Springer}
}

@article{shapiro2025hybrid,
  title={Hybrid Quantum-Classical Machine Learning with PennyLane: A Comprehensive Guide for Computational Research},
  author={Shapiro, Sidney},
  journal={arXiv preprint arXiv:2511.14786},
  year={2025}
}

@article{xu2023resnet,
  title={ResNet and its application to medical image processing: Research progress and challenges},
  author={Xu, Wanni and Fu, You-Lei and Zhu, Dongmei},
  journal={Computer Methods and Programs in Biomedicine},
  volume={240},
  pages={107660},
  year={2023},
  publisher={Elsevier}
}

@inproceedings{chen2024quantum,
  title={Quantum-classical-quantum workflow in quantum-hpc middleware with gpu acceleration},
  author={Chen, Kuan-Cheng and Li, Xiaoren and Xu, Xiaotian and Wang, Yun-Yuan and Liu, Chen-Yu},
  booktitle={2024 International Conference on Quantum Communications, Networking, and Computing (QCNC)},
  pages={304--311},
  year={2024},
  organization={IEEE}
}

@article{cai2022theoretical,
  title={Theoretical foundations of t-sne for visualizing high-dimensional clustered data},
  author={Cai, T Tony and Ma, Rong},
  journal={Journal of Machine Learning Research},
  volume={23},
  number={301},
  pages={1--54},
  year={2022}
}

\end{document}